\documentclass[conference]{IEEEtran}
\IEEEoverridecommandlockouts
\usepackage{generic}
\usepackage{cite}
\usepackage{xcolor}
\usepackage{amsmath,amssymb,amsfonts}
\usepackage{algorithmic}
\usepackage{graphicx}
\usepackage{algorithm,algorithmic}
\usepackage{hyperref}
\usepackage{balance}
\usepackage{textcomp}
\def\BibTeX{{\rm B\kern-.05em{\sc i\kern-.025em b}\kern-.08em
    T\kern-.1667em\lower.7ex\hbox{E}\kern-.125emX}}
\makeatletter
\newcommand{\linebreakand}{%
  \end{@IEEEauthorhalign}
  \hfill\mbox{}\par
  \mbox{}\hfill\begin{@IEEEauthorhalign}
}
\makeatother
\begin{document}

\title{Sleep Position Classification using Transfer Learning for Bed-based Pressure Sensors}

\author{\IEEEauthorblockN{Olivier Papillon}
\IEEEauthorblockA{\textit{Systems and Computer Engineering} \\
\textit{Carleton University}\\
Ottawa, Canada \\
olypapillon@cmail.carleton.ca}
\and
\IEEEauthorblockN{Rafik Goubran, Life Fellow IEEE}
\IEEEauthorblockA{\textit{Systems and Computer Engineering} \\
\textit{Carleton University}\\
\textit{Bruyère Health Research Institute}\\
Ottawa, Canada \\
goubran@sce.carleton.ca}
\and
\IEEEauthorblockN{James Green, Senior Member IEEE}
\IEEEauthorblockA{\textit{Systems and Computer Engineering} \\
\textit{Carleton University}\\
Ottawa, Canada \\
jrgreen@sce.carleton.ca}
\and
\IEEEauthorblockN{Julien Larivière-Chartier}
\IEEEauthorblockA{\textit{Bruyère Health Research Institute} \\
\textit{Systems and Computer Engineering} \\
\textit{Carleton University}\\
Ottawa, Canada \\
JLariviereChartier@bruyere.org}
\and
\IEEEauthorblockN{Caitlin Higginson}
\IEEEauthorblockA{\textit{Sleep Research Unit} \\
\textit{University of Ottawa Institute for Mental} \\
\textit{Health Research at the Royal}\\
Ottawa, Canada \\
Caitlin.Higginson@theroyal.ca}
\and
\IEEEauthorblockN{Frank Knoefel, MD}
\IEEEauthorblockA{\textit{Systems and Computer Engineering} \\
\textit{Carleton University}\\
\textit{Bruyère Health Research Institute}\\
Ottawa, Canada \\
fknoefel@bruyere.org}
\linebreakand
\IEEEauthorblockN{Rébecca Robillard}
\IEEEauthorblockA{\textit{School of Psychology,} \\
\textit{University of Ottawa} \\
\textit{University of Ottawa Institute for Mental} \\
\textit{Health Research at the Royal}\\
Ottawa, Canada \\
Rebecca.Robillard@uottawa.ca}
}

\maketitle
\begin{abstract}
Bed-based pressure-sensitive mats (PSMs) offer a non-intrusive way of monitoring patients during sleep. We focus on four-way sleep position classification using data collected from a PSM placed under a mattress in a sleep clinic. Sleep positions can affect sleep quality and the prevalence of sleep disorders, such as apnea. Measurements were performed on patients with suspected sleep disorders referred for assessments at a sleep clinic. Training deep learning models can be challenging in clinical settings due to the need for large amounts of labeled data. To overcome the shortage of labeled training data, we utilize transfer learning to adapt pre-trained deep learning models to accurately estimate sleep positions from a low-resolution PSM dataset collected in a polysomnography sleep lab. Our approach leverages Vision Transformer models pre-trained on ImageNet using masked autoencoding (ViTMAE) and a pre-trained model for human pose estimation (ViTPose). These approaches outperform previous work from PSM-based sleep pose classification using deep learning (TCN) as well as traditional machine learning models (SVM, XGBoost, Random Forest) that use engineered features. We evaluate the performance of sleep position classification from 112 nights of patient recordings and validate it on a higher resolution 13-patient dataset. Despite the challenges of differentiating between sleep positions from low-resolution PSM data, our approach shows promise for real-world deployment in clinical settings.
\end{abstract}

\begin{IEEEkeywords}
pressure sensors; ViT; VITMAE; VitPose; CNN; sleep position; health monitoring; transfer learning; classification; deep learning; pre-training
\end{IEEEkeywords}

\section{Introduction}
With age comes a greater reliance on healthcare services as age-related health conditions become more prevalent. Remote health monitoring could reduce the strain on the Canadian healthcare system. \cite{b1} Polysomnography, or the study of sleep, becomes increasingly important as the prevalence of sleep-related disorders increases with age. Researchers have proposed using bed-based pressure-sensitive mats (PSM) placed under the mattress as an unobtrusive method of monitoring patients. To this end, bed-based PSM have been installed at the Royal Ottawa Hospital sleep study lab, where they record time-varying contact pressure data throughout the night. This technology has proven useful for monitoring bed-based mobility \cite{b34}, sleep stages \cite{b3,b5,b6,b7,b23,b30}, pose estimation \cite{b6,b7,b10,b15,b23,b28}, and respiration rate\cite{b4,b29}. 


In the present study, we focus on automatically recognizing body positions from data recorded by the PSM. Research has indicated that an individual's sleep position significantly influences their sleep quality, and the prevalence and severity of sleep disorders such as sleep apnea \cite{b13}.
As such, we are interested in accurately and automatically recognizing the body positions of sleeping patients using only PSM data. Accurate pose estimation enables researchers to select the most appropriate estimation approaches for downstream tasks, such as respiration rate estimation or patient movement detection. For example, robustly estimating respiration rate from PSM data requires different approaches for side-lying subjects vs. supine; by providing the ability to determine a patient’s pose reliably, we improve downstream respiration rate estimation and achieve clinical benefit.

To do this, we use machine learning (ML) to analyze PSM data and extract meaningful patterns and valuable insights about the data. Recent research has extracted sleep positions from PSM data using ML algorithms such as Support Vector Machines (SVM) and various Deep Learning (DL) architectures \cite{b3},\cite{b5} -\cite{b7}. These methods are effective; however, recent DL architectures have demonstrated improved classification and computer vision task performance. Since PSM data represent images over time (i.e., video), the use of DL is expected to outperform traditional ML that requires hand-engineered feature extraction. Although DL methods can be very powerful, training DL models still poses a significant challenge due to the large amount of labeled data required for training \cite{b22}. To overcome the present shortage of labeled training data, we can utilize transfer learning to leverage state-of-the-art pre-trained DL models and fine-tune them for our specific task. This research makes contributions to the fields of computer vision, machine learning, and medical measurements and instrumentation by examining the effectiveness of newer DL models to detect sleeping poses from PSM data. Given that PSMs represent a privacy-preserving sensor modality, such a system could ultimately be deployed in a real-time sleep lab or for remote monitoring at home and long-term care facilities.

\section{Related Works}
\subsection{Posture Recognition using Bed-Based Pressure Sensor Arrays}
In \cite{b7}, Foubert \textit{et al.} performed sleep position classification using classical ML methods to classify the sleep positions. The study used bed-based PSM with 44 sensors and 25 participants of various ages. The researchers extracted features from the PSM data (e.g., the sum of sensor values, the center of pressure) to train SVM and K-NN classifiers to recognize six different sleeping poses: Prone, Supine, Left Lateral, Right Lateral, Left Fetal, and Right Fetal. In the present study, this method served as a baseline for comparison of the effectiveness of the newer pre-trained DL architectures.



\subsection{Sleep-Wake and Body Position Classification with Deep Learning using Pressure Sensor Mat Measurements}
In \cite{b6}, Green \textit{et al.} performed sleep-wake and body position estimation using a TCN (Temporal Convolution Network) with PSM data similar to those used in the present study. PSM data and gold standard body positions were annotated at the Royal Ottawa Hospital. A low-resolution PSM was used that comprised only 72 sensors. Only three body positions were included in the dataset since the Supine and Prone positions were merged into a single position to simplify the training of the classifiers. The present study expands on this study by doubling the number of sensors (144 sensors) and treated the problem as a 4-way classification rather than 3-way. By including Prone as a distinct body position, the problem becomes substantially more difficult since this position represents a minority class and is difficult to distinguish from other poses based on PSM data alone. 


\subsection{A Pressure Map Dataset for In-bed Posture Classification}
In \cite{b15}, Pouyan \textit{et al.} presented the publicly available Pmat dataset. This dataset was collected from 13 participants laying on a Force Sensitive Application pressure mapping mattress comprising a 32 × 64 grid of sensels (2048 total) with approximately 1-inch separation. The dataset included pressure maps for adults in 17 different postural poses, which were categorized into three classes: Supine, Right, and Left. This dataset has been used by a variety of research groups for pose estimation, sleeper identification, and other tasks. Of note, Doan \textit{et al.} \cite{b32} and Davoodnia and Etemad  \cite{b31} presented leave-one-subject-out results for 3-way sleep position estimation from PSM exceeding 95\% accuracy. In \cite{b31}, Davodnia and Etemad present simultaneous identity and posture recognition in smart beds using deep multitask learning techniques. In \cite{b32}, Doan {et al.} used an EfficentNet B0 based classifier to identify 17 sleeping postures as patients rested in the hospital beds. The research highlighted the potential of real-time posture recognition systems in enhancing patient care and optimizing bed utilization in hospital settings.

In the current study, we use the Pouyan \textit{et al.}  dataset \cite{b15} to evaluate the effectiveness of our proposed approach on higher-resolution PSM images, and to demonstrate that our proposed approach generalizes beyond the dataset on which it is developed. We further show that, similar to the results observed in \cite{b10} and \cite{b35}, we can achieve performance gains on our clinical dataset by pre-training models on the larger research dataset, despite the difference in PSM resolution.

\subsection{ViTMAE}
In \cite{b11}, He \textit{et al.} used masked autoencoding (MAE) to accomplish self-supervised learning for pre-training of a ViT model. During MAE, randomly selected image patches were masked and the ViT was trained to reconstruct the missing pixels using self-supervised learning. He \textit{et al.} showed that using MAE for pre-training, followed by fine-tuning on a downstream task, led to improved generalization and robustness in scenarios with limited application-specific labelled data. Such pre-training followed by fine-tuning on an application-specific dataset represents a form of transfer learning. In the present study, we employ a ViT that is pre-trained from ImageNet using MAE and fine-tuned on our clinical PSM dataset. 

\subsection{ViTPose}
In \cite{b14}, Xu \textit{et al.} used ViTPose, a model leveraging the ViT architecture for human pose estimation (keypoint detection). Unlike traditional methods that utilized convolutional layers, ViTPose directly employed a ViT to predict human joint positions, using its global receptive field and robust representational abilities. The researchers demonstrated that ViTPose, when pre-trained on large datasets and fine-tuned on pose-specific tasks, achieved superior accuracy and efficiency compared to state-of-the-art models. This approach illustrated the potential of transformers in pose estimation tasks. In our work, we used the pre-trained ViTPose model and fine-tune it for our task of pose classification on a PSM dataset.

\subsection{Developing a Pressure Sensitive Mat using Proximity Sensors for Vital Sign Monitoring}
In \cite{b34}, Selzler \textit{et al.} presented the design and development of a novel flexible pressure-sensitive mat (PSM) for non-intrusive patient monitoring. This system utilized infrared proximity sensors arranged in a matrix on a flexible printed circuit board, covered with a silicone rubber layer. The mat was designed to measure heart and respiratory rates by analyzing pressure distribution changes. 
\subsection{PSM-Video Fusion for Pose Estimation}
In \cite{b23}, Kyrollos \textit{et al.} use multi-modal registration techniques to spatially align overhead video with PSM data. A perspective transform is used to align the modalities using several registration points captured during data collection. Later, in \cite{b10}, they show that ViT can effectively fuse video and PSM information without explicit registration. Interestingly, they show that PSM data are more informative for pose estimation than overhead depth video, when the patient is covered by a blanket.

\section{Methods}

\subsection{Clinical Data Collection}
\begin{figure}[h!]
    \centering
    \resizebox{9.5cm}{!}{\includegraphics{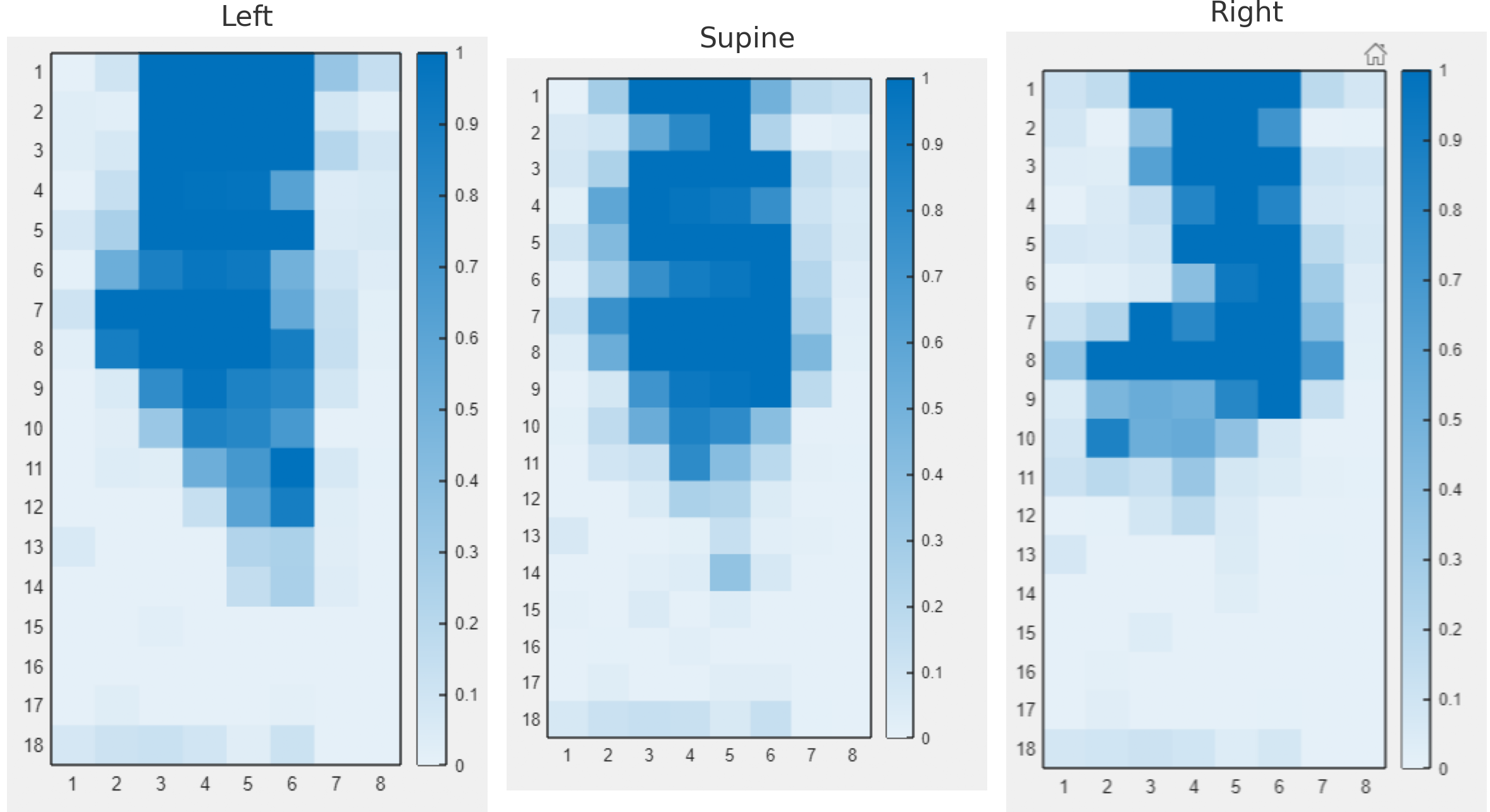}}
    \caption{Output image of the 18x8 PSM of a patient in three different poses sleeping on the bed. (a) Left Lateral Decubitus. (b) Supine. (c) Right Lateral Decubitus.}
    \label{fig:psm-image}
\end{figure}

During an overnight study at the Royal Ottawa Hospital's Sleep Disorder Clinic, consenting patients were monitored in two rooms equipped with PSMs. As patients slept, PSM data were collected. These data were later annotated by a technician with sleep position (4 poses) and sleep stages (not used in the present study).
Each bed in the clinic is outfitted with several PSMs created by Tactex Control Inc. These mats consist of a 3x8 grid of fiber optic proximity sensors that detect pressure at a sampling rate of 10 Hz. These sensors output values ranging from 0 to 2046, which were normalized to a scale of 0 to 1.
Each room has three PSMs connected to a data acquisition box (DAQ) for the upper body and three PSMs for the mid-to-lower body, connected to another DAQ. This configuration produces an 18x8 low-resolution image output, as shown in Fig.\ref{fig:psm-image}, with a total of 144 sensels. The data collected is saved to SD cards connected to each DAQ, before data are transferred to a workstation for subsequent analysis.


In one room, the mats are spaced evenly on a flat metal surface, while in the other room, an Evacu-Sled is used, necessitating an uneven surface with spaced-out mats. To ensure accurate data collection, a wood plank creates an even, flat surface for the PSMs to sense pressure correctly. 


\subsection{Time Synchronization of Sensor Data}
As mentioned previously, each room has two individual DAQs that continuously collect data throughout the night from 8PM-8AM the next day. Data are collected independently from the upper and lower PSM sections and there is often a small time difference between each data collection system. This is due to each DAQ having its own internal real-time clock that is synced to a common network time after each data backup. However, when the data are recorded, each DAQ internal clock drifts slightly, resulting in time alignment issues between the upper and lower-body PSMs. Throughout a single night, the shift in the clock is insignificant, but it accumulates over longer periods. To analyze the data from the PSM, we must synchronize the upper and lower-body PSMs and then synchronize them to the time used by the technician when annotating gold-standard sleep positions in the sleep event logs.

To synchronize the top and bottom mats, we use a common ``calibration'' event that can be seen on both mats and appears on both the upper and lower PSM sections. The patient will typically enter the bed between 9-11 pm and leave the bed around 5-6 am. This event can be detected by looking for the maximum non-zero pressure on both mats when the overall centre of pressure indicates that the patient is visible on both PSM sections. The difference in the recorded time of the bed entry event on each PSM section is used to time-align the upper and lower-body PSMs.

\subsection{Gold Standard Data Annotation}
To correlate the hand-annotated sleep position and stage information created by the technician (which is based on patient observations and not on the PSM data),  we must also synchronize the expert annotations in the sleep logs with the PSM data from the upper-body and lower-body PSMs. However, the time from the PSM data does not align exactly with the video-annotated logs, necessitating a synchronization event that can be reliably observed on both the PSMs and the logs. For this we use the biocalibrations from the sleep technician, which includes guiding the patient through various respiratory maneuvers, such as taking a big breath in and out, holding breath, and breathing through the nose and mouth. Although these are primarily used to verify the correct functioning of all instrumentation, they can also be seen in the PSM data. We employ this event to determine the time difference between the mats and the sleep logs to synchronize them. This synchronization allows us to extract the gold standard sleep positions from the logs and obtain a dataset that contains the PSM data aligned with the sleep position annotations. 


The dataset comprises four sleep positions: Left Lateral Decubitus (Left), Right Lateral Decubitus (Right), Supine, and Prone. 
The Prone position has been added for this study and represents a minority class in the dataset that is challenging to differentiate from the Supine position in the PSM data. 


\subsection{Preprocessing}
Our dataset consists of 112 nights of patient data recordings, where a typical patient will sleep an average of 7 hours. Data are recorded at 10 Hz, thus each night comprises approximately 250,000 samples. We divided our dataset using k-fold cross-validation with 5 folds. Each fold consists of approximately 22 patients, resulting in a training set with 90 patients and a testing set with 22 patients in each fold. Data from the same patient are never used to both train and test a classifier, enabling us to detect and avoid overfitting. Considering that patients maintain the same position for extended periods, we downsampled our PSM data to 1 Hz. This allowed us to reduce the number of samples to 25k per night, per patient, while losing minimal information. We also excluded transient positions, which refer to the moments when the patient is switching between positions. To reduce compute times, we also eliminated nearly identical consecutive frames. 

This was achieved by identifying the first difference between consecutive frames and subsequently removing those frames that fell below a minimum threshold. This process allowed us to remove duplicates, resulting in a total of 560k samples for the training set and 140k samples for the testing set in each fold.

\subsection{Machine Learning Methods}
In addition to the pre-trained ViT models (ViTMAE and ViTPose), we include several baseline models for comparison, including TCN and traditional machine learning approaches using engineered features.

The ViTMAE model is a state-of-the-art model, pre-trained on the ImageNet dataset, which consists of RGB images with 1000 object classes and a size of 224x224 pixels. However, the data produced by the PSM system is normalized pressure data with only one channel and only 18x8 pixels. We use bilinear interpolation to upscale the images to square 18x18 pixels. To make it compatible with the model, we modify the positional embeddings of the transformer to support a smaller image input size, and the three-channel RGB input weights are summed into one channel to accommodate the one-channel PSM data.

The ViTPose model, a state-of-the-art model, incorporates a pre-trained ViTMAE as its backbone and is fine-tuned for keypoint detection in human pose estimation. It processes RGB images of size 224x224. For application to PSM data, we adjust the model input layer similarly to VitMAE except that the weights are randomized. Furthermore, since ViTPose is configured for keypoint detection, modifications are required to perform classification tasks. To ensure compatibility with PSM data, we replace the transformer's existing head with a linear layer that maps the transformer's output to the number of classes.

A TCN model, similar to the one used by Green \textit{et al.} \cite{b6}, where they use one TCN block with 128 filters of size 15, which is then passed to a classification head that consists of a fully connected network of 32 neurons, followed by a Relu activation layer, into a linear projection to the number of classes. We use the interpolated 18x18 images for the TCN network as well.


We also used the PSM-engineered features developed by Foubert \textit{et al.} \cite{b7} in combination with various ML model architectures. These features have been shown to work well with an SVM model. Additionally, we also employed two other model architectures, a gradient-boosted decision tree model (XGBoost) and a random forest classifier.

\subsection{External PSM Dataset}
To further asses the effectiveness of VITMAE and VitPose, we evaluate our proposed method on a second, higher-resolution PSM dataset from \cite{b15}, where each PSM frame comprises 32x64 pixels.  We employ zero-padding to interpolate the images to a resolution of 64 × 64 pixels since our proposed approach requires square input images. Additionally, we used this higher-resolution PSM dataset to pre-train the ViTMAE model and then finetune for our clinical dataset.


\subsection{Model Development}
ViTMAE, ViTPose, and TCN models were run on an A100 GPU using Pytorch and the Timm library in Python. A hyperparameter sweep of the learning rate, weight decay, and epochs was completed to find the models with the highest average 4-way accuracy over the five folds. For the SVM, XGBoost, and Random Forest classifier, we used sklearn and the xgboost libraries in Python; hyperparameter sweeping was performed using the GridSearchCV model selection utility from the scikit-learn library.

\section{Results and Discussion}

We compared different ML architectures for sleep position classification from PSM data. Four-way classification accuracy and the macro-averaged F1 score were averaged over the five folds of 22 test patients in each fold, the results are shown in Table \ref{tab:Results}. Although K-fold cross-validation ensures that the same data are never used for both training and testing a model, repeated cross-validation during hyperparameter optimization can sometimes lead to overfitting.

\begin{table}
\centering
\caption{\textbf{Comparison of results for the sleep position classification for our clinical dataset}}
\begin{tabular}{cllclc}
         Method & & Accuracy & F1 (macro)\\ \hline
         ViTMAE (pre-trained on \cite{b15}) & & \textbf{0.770}& \textbf{0.731}\\ 
         ViTMAE & & 0.643& 0.598\\ 
         ViTPose & & 0.649 & 0.583\\ 
         TCN & & 0.398 & 0.236\\ 
         SVM & & 0.352 & 0.167\\ 
         Random Forest & & 0.322 & 0.252\\ 
         XGBoost & & 0.324 & 0.251\\  
    \end{tabular}
    \label{tab:Results}
\end{table}



\begin{table}
\centering
\caption{\textbf{Comparison of results posture classification using external test dataset}}
\begin{tabular}{cllclc}

         Method & &  Accuracy &  F1 (macro)\\ \hline  
          ViTMAE & &  0.994&  0.951\\ 
          ViTPose & &  0.970 &  0.967\\ Multitask CNN \cite{b31}   & &  0.992  &   - \\
          EfficientNet \cite{b32} & &  0.953  &  - \\ 
    \end{tabular}
    \label{tab:High-Res}
\end{table}

The results indicated that the pre-trained ViT models outperformed all the other methods by a wide margin. ViT performed the best when it had additional pre-training from \cite{b15} and ViTPose performed very similarly to ViTMAE when only pre-trained from ImageNet; their main difference was a tradeoff between F1 and Accuracy. TCN were the third-most performant classifier. The dominance of the ViT models indicated that we can improve pose classification by using pre-trained models, even when there exists a domain shift between the pre-training task (RGB image classification, pose estimation, high-resolution pressure images) and the target domain (single-channel low-resolution PSM data). However, we noticed that the overall accuracy of all models was relatively low. This reflected the difficulty of the task, partly attributable to the low resolution of the PSM itself.
\begin{figure}[htbp]
    \centering
    \resizebox{6cm}{!}{\includegraphics{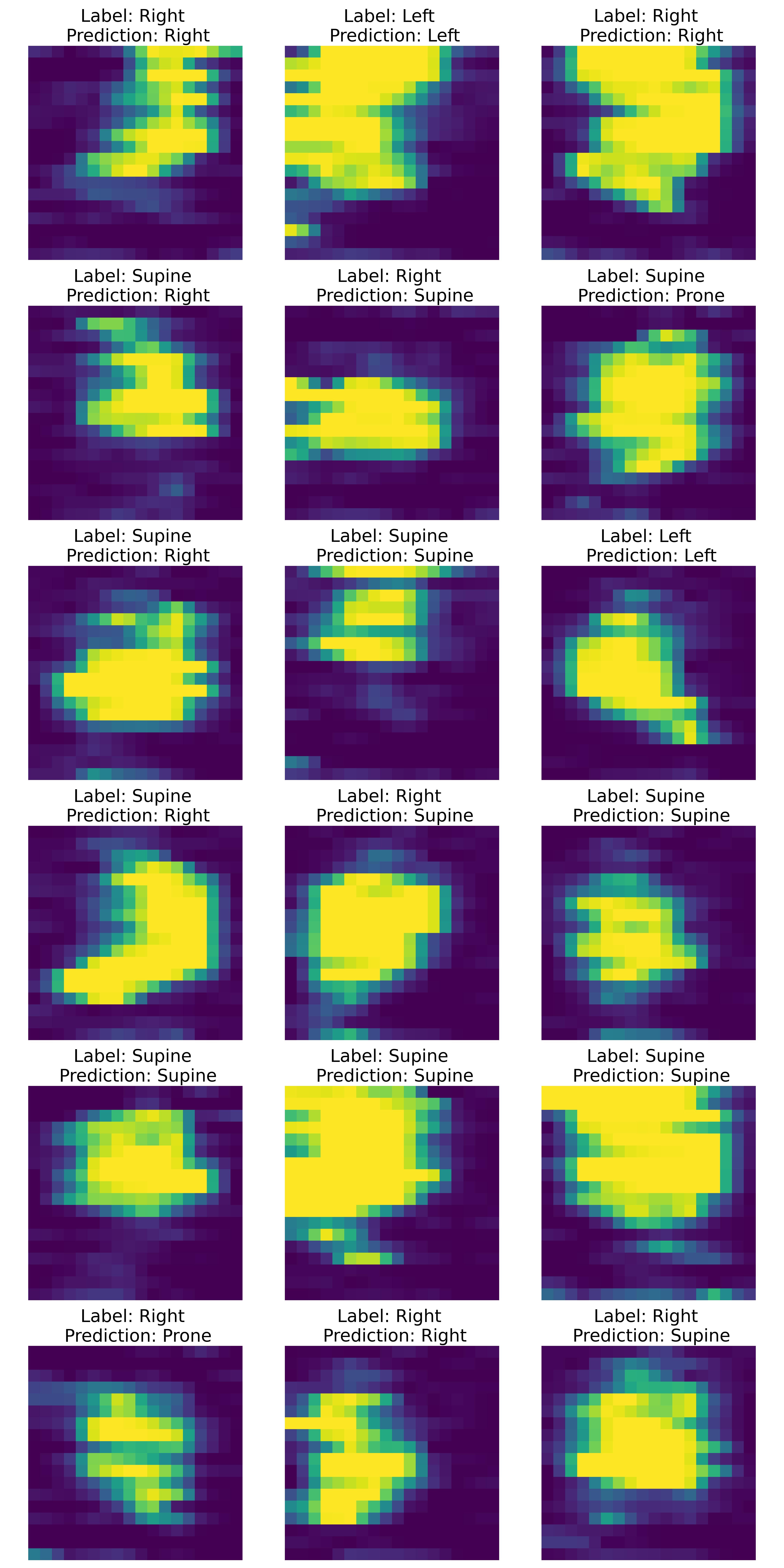}}
    \caption{Examples comparing the actual class vs. the class predicted by ViTMAE from PSM images}
    \label{fig:pred-image}
\end{figure}

The task difficulty due to low sensor resolution is illustrated in Fig. \ref{fig:pred-image}, where we compared the ViTMAE model's predictions to the true labels for several example PSM images. The errors made by the model were similar to the mistakes expected of a human observer when attempting to differentiate between sleep positions. Sleep positions could be challenging to distinguish because they often appeared similar on the PSMs. For example, in the fourth case where the label was Supine, the pressure image appeared similar to the Right position and could be easily mistaken.

\subsection{Validation of Proposed Methods an External High-Res PSM Dataset}
We have validated our proposed method on higher-resolution PSM data from \cite{b15}. Pose classification accuracy increases from 0.643 to 0.994 for VitMAE, and from 0.649 to 0.970 for ViTPose, as shown in Table \ref{tab:High-Res}. These results are competitive state-of-the-art methods developed specifically for this dataset, when evaluated using a LOSO test protocol \cite{b31,b32}. This performance boost is likely attributable to the increased resolution of the PSM images (32 x 64 instead of 18x8 sensels). Furthermore, the dataset from \cite{b15} is a research dataset, recorded under optimal conditions, whereas our dataset is a true clinical dataset with actual patient data recorded under more natural and varied conditions.

Our method is the first reported ViT-based approach for pose estimation from PSM, and the first method to be developed and validated using actual clinical data. Ultimately, the results on the dataset from \cite{b15} serve to validate our proposed approach and demonstrate its ability to generalize to a new dataset and to higher-resolution PSM data. Additionally, as shown in Table I, when the external dataset is used to pre-train our ViTMAE model, substantial improvements in accuracy are observed on our clinical dataset. This further illustrates the ability of ViT models to benefit from pre-training, even when the pre-training task differs from the downstream classification task. (here, in terms of PSM resolution)

\section{Conclusion}
Our study has demonstrated that pre-trained DL transformer models can achieve promising performance on pose classification from clinical PSM data through fine-tuning. Data were collected from actual patients in a clinical overnight sleep study lab using low-resolution PSM sensors. Although this dataset proved to be highly challenging, our proposed approach outperformed ML approaches based on engineered features. Relative to the performance on the research dataset from \cite{b15}, the somewhat diminished pose classification accuracy on our clinical sleep clinic dataset may be explained by the complexity of working with PSM data collected in actual clinical settings, rather than research subjects in controlled lab conditions, and the relatively low-density PSM sensors used in the present study (8 * 18 sensels vs. 64 * 27 sensels used in other studies). When validating the proposed approach on a secondary dataset with higher-resolution PSM data the method generalized well, achieving comparable accuracy to the state of the art. The present results indicate that a suitably trained ViT holds the potential for practical sleep pose estimation in the clinic.

In future work, we intend to improve the model's performance by accounting for class imbalance. We will also investigate combining PSM-based pose estimation with PSM-based vital sign estimation, such as respiratory rate estimation and apnea detection, such that the estimated pose provides context information to the vital sign estimator. 

\section*{Acknowledgment}
We acknowledge the support of the Natural Sciences and Engineering Research Council of Canada (NSERC).

 F. Knoefel acknowledges the support for the Bruyère Health Research Institute Chair in Research in Technology for Aging in Place.

This study followed IRB-approved protocol  \#112123 ``Sleeping and Health Improvements; Following Longitudinal Trajectories (SHIFT)''
\balance

\vspace{12pt}

\end{document}